\documentclass[lettersize,journal]{IEEEtran}
\usepackage{amsmath,amsfonts}
\usepackage{algorithmic}
\usepackage{algorithm}
\usepackage{array}
\usepackage[caption=false,font=normalsize,labelfont=sf,textfont=sf]{subfig}
\usepackage{textcomp}
\usepackage{stfloats}
\usepackage{url}
\usepackage[dvipsnames]{xcolor}
\usepackage{hyperref}
\usepackage{verbatim}
\usepackage{graphicx}
\usepackage{cite}
\usepackage{multirow}
\usepackage[flushleft]{threeparttable}
\begin{document}

\title{MorphGANFormer: Transformer-based Face Morphing and De-Morphing}
\author{Na Zhang, Xudong Liu, Xin~Li,~\IEEEmembership{Fellow,~IEEE}, Guo-Jun Qi
}



\maketitle

\begin{abstract}
Semantic face image manipulation has received increasing attention in recent years. StyleGAN-based approaches to face morphing are among the leading techniques; however, they often suffer from noticeable blurring and artifacts as a result of the uniform attention in the latent feature space. In this paper, we propose to develop a transformer-based alternative to face morphing and demonstrate its superiority to StyleGAN-based methods. Our contributions are threefold. First, inspired by GANformer, we introduce a bipartite structure to exploit long-range interactions in face images for iterative propagation of information from latent variables to salient facial features. Special loss functions are designed to support the optimization of face morphing. Second, we extend the study of transformer-based face morphing to demorphing by presenting an effective defense strategy with access to a reference image using the same generator of MorphGANFormer. Such demorphing is conceptually similar to unmixing of hyperspectral images but operates in the latent (instead of pixel) space. Third, for the first time, we address a fundamental issue of vulnerability-detectability trade-off for face morphing studies. It is argued that neither doppelganger nor random pair selection is optimal, and a Lagrangian multiplier-based approach should be used to achieve an improved trade-off between recognition vulnerability and attack detectability. 
\end{abstract}

\begin{IEEEkeywords}
transformer, face morphing, De-morphing.
\end{IEEEkeywords}

\section{Introduction}
\label{sec:intro}
\par
\begin{figure*}
\centering
\includegraphics[width=1.0\linewidth]{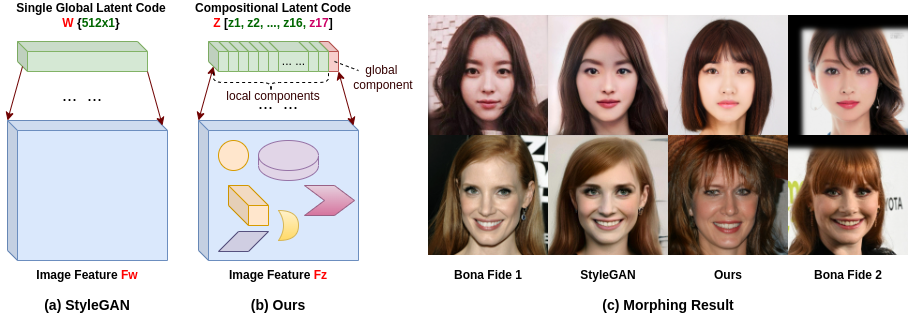}
\caption{Illustration of latent code modulation of (a) StyleGAN and (b) Our MorphGANFormer. StyleGAN uses a single global-style latent code to modulate the whole scene uniformly in one direction. Ours is a compositional latent code with 16 local- and one-global-style components to impact different regions in the image allowing for spatially finer control over the generation process bidirectionally. Figure (c) shows some morphing results of StyleGAN-based model and our MorphGANFormer (ours contain fewer visual artifacts).}
\vspace{-0.2in}
\label{fig:teaser}
\end{figure*}

 With the rapid development of deep-learning technology, automatic face recognition (FR) has become a key method in security-sensitive applications of identity management (e.g. travel documents). However, the face recognition system (FRS) is vulnerable to face morphing attacks \cite{venkatesh2021face}, which aim to create facial images that can be successfully matched to more than one person. Existing face-morphing methods can be classified into two categories. One is performed on the image level via landmark interpolation, like OpenCV \cite{opencv}, FaceMorpher \cite{facemorpher}, LMA \cite{damer2018morgan}, WebMorph \cite{webmorph}. The other works are performed by manipulating latent codes of generative adversarial networks (GAN), such as MIPGAN-II \cite{zhang2021mipgan}, MorGAN \cite{damer2018morgan}, StyleGAN \cite{karras2020analyzing}. Both approaches have serious limitations. For landmark-based methods, as the morphing process translates landmarks and the associated texture, misaligned pixels tend to generate artifacts and ghost-like images, making the images unrealistic (i.e., easy for a human observer to detect). Similarly, for GAN-based methods, unpleasant visual artifacts, such as noticeable blurring and abnormal image patterns, often occur, often making morphed faces unnatural (see Fig. \ref{fig:teaser}). It is natural to seek an alternative approach to face morphing attacks.

\par Transformer-based architectures have found successful applications in natural language processing \cite{vaswani2017attention, dai2019transformer, devlin2018bert}, object detection \cite{carion2020end}, image restoration \cite{chen2021pre, liang2021swinir}, video inpainting \cite{liu2021fuseformer, zeng2020learning}, image synthesis \cite{zhang2022styleswin, hudson2021generative, arad2021compositional, zhao2021improved, liu2021swin, jiang2021transgan}, and so on. Inspired by the capability of exploiting the long-range dependency of GANformer \cite{hudson2021generative}, we propose to develop the GANformer-based morphing attack in a compositional latent space, as shown in Fig. \ref{fig:teaser} (b). The compositional latent space is composed of multiple latent components in local-style and one latent component in global-style, respectively. Such a compositional design allows us to have finer control of salient regions (e.g., face in the foreground) than the less important region (e.g., background). Meanwhile, MorphGANFormer is bidirectional, allowing the propagation of information between latent codes and image features in both directions. In addition to long-range dependency, duplex attention on bipartite graphs facilitates the synthesis of high-resolution by keeping computation linear. 

\par Under the transformer-based framework, we focus on the design of latent code in the compositional space. Unlike GANformer \cite{hudson2021generative} which simply adopts the loss function of StyleGAN studies \cite{karras2019style,karras2020analyzing}, we have designed a class of loss functions specifically tailored for face morphing applications. Our design attempts to expedite the search for a suitable latent code by combining the strengths of both landmark-based and GAN-based approaches. Both facial landmarks and features (e.g., histogram of orientated gradients \cite{dalal2005histograms}) are included as content-related regularization terms. Style-related regularization consists of VGG-based perceptual loss and pixel-based MSE loss. The tradeoff between the style and context loss terms allows us to strike an improved balance between visual quality (i.e., fewer artifacts) and attack success (i.e., better matching).

Like other security systems, morphing attacks and defenses co-evolve in a never-ending race. Morphing and demorphing \cite{ferrara2017face,ferrara2018face} are two sides of the same coin, although relatively less attention has been paid to demorphing studies in the literature. The other contribution of this work is to conduct a dual study of demorphing in latent space, which complements our construction of MorphGANFormer. For the first time, we address a fundamental issue of vulnerability-detectability tradeoff for face morphing studies - i.e., what pair of images should be used in morphing study? A pair of similar images (e.g., doppelganger \cite{rottcher2020finding}) might be desirable from a recognition vulnerability perspective but suffers from being more easily detectable (i.e., higher APCER/BPCER rates). On the other hand, two random faces enjoy the advantage from the attack detectability perspective, but sacrifice the recognition vulnerability (i.e., lower MMPMR rate \cite{scherhag2017biometric}). It is argued that neither the selection of doppelgangers nor random pairs is optimal and a Lagrangian multiplier-based approach should be used to achieve an improved trade-off between the recognition vulnerability and the detectability of the attack \cite{damer2018morgan}. The main contributions of this paper are summarized below.

\begin{itemize}
\item Design a transformer-based GAN model with a compositional latent space. It is made up of 16 local-style latent code components and one extra global-style component with $32\times 1$ dimension for each. Different components can impact different regions in the image, allowing for spatially finer control over the generation process bidirectionally.

\item Design special loss functions to improve the performance of the latent code optimization problem by maximizing the similarity between the generated face and the target face. Four types of loss function are adopted: biometric loss, landmark-based loss, perceptual loss, and pixel-wise mean square error (MSE).

\item Extend the study of transformer-based face morphing to demorphing using the same generator. With the final morphed face and a given trusted live capture of one bona fide face, we have shown how to successfully restore the other bona fide face.

\item Experimental results with both Doppelganger and random selection to demonstrate the trade-off between recognition vulnerability and attack detectability. We hope that this line of research will lead to a deeper understanding of adversarial attack and defense in the study of face morphing and demorphing.
\end{itemize}

\section{Related Works}
\label{sec:work}
\subsection{Landmark-based Generation} 
\par Morphed face is initially performed by detecting facial landmarks of two bona fide faces. The final morphed face is generated by landmark interpolation and texture blending. The landmark-based method, as the name suggests, works by obtaining landmark points on facial regions, like the nose, eyes, mouth, etc. The landmarks obtained from two bona fide faces are warped by moving the pixels to different, more averaged positions. There exist different procedures for warping in the literature. Delaunay triangulation is a popular one. The basic idea is to perform Delaunay Triangulation on the three sets of landmarks (2 bona fides and their average points) and do affine transform and warping. The two warped faces will do alpha blending, and then the final morphed face is generated.

\par The most popular methods contain OpenCV \cite{opencv}, FaceMorpher \cite{facemorpher}, LMA \cite{damer2018morgan}, WebMorph \cite{webmorph}, etc. In the OpenCV \cite{opencv} algorithm, the landmarks of the bona fide faces are obtained by Dlib \cite{king2009dlib} and then used to form Delaunay triangles \cite{lee1980two}, which in turn are warped and mixed with alpha.  FaceMorpher \cite{facemorpher} is also an open-source tool similar to OpenCV, but with the STASM \cite{milborrow2014active} landmark detector instead. Both algorithms create morphs with noticeable ghosting artifacts, as the region outside the area covered by these landmarks is simply averaged. WebMorph \cite{webmorph} is an online landmark-based morphing tool that requires 189 landmarks, to generate morphed images with better alignment and of higher visual quality. Ghosting artifacts are still visible and prominent around the hair and neck area. Similar to OpenCV and FaceMorpher, LMA \cite{damer2018morgan} is performed by detecting facial landmarks, the mean face points for each image are calculated and each image is then warped to sit on these coordinates after performing the Delaunay triangulation, but uses 194 points detected by an ensemble of randomized regression trees \cite{kazemi2014one}. One special is a combined private Morphs tool used in the AMSL face morph image database \cite{amsl}. This tool can generate very realistic morphs with virtually no ghosting artifacts, even around the hair and neck area, thanks to the additional Poisson image editing.

\subsection{GAN-based Generation} 
\par GAN-based model has made a major breakthrough in high-quality image synthesis, especially on human faces \cite{karras2019style}. Taking advantage of the advanced GAN architectures and their ability to produce synthetic images, we proposed a few GAN-based morphing approaches that avoid image-level interpolation. It works by embedding the images into the intermediate latent space. First, two bona fide face images are mapped into the latent space to obtain their latent codes. And then a linear combination of two latent codes is made to obtain a final latent code, which is put into the generator of the pre-trained GAN model to synthesize the morphed image. 

\par StyleGAN2 \cite{karras2020analyzing} is a morphing algorithm that can generate realistic high-resolution faces. Based on StyleGAN \cite{karras2019style}, the MIPGAN-II \cite{zhang2021mipgan} was designed to generate images with higher identity preservation by introducing a loss to optimize identity preservation in the latent vector. MorGAN \cite{damer2018morgan} is based on automatic image generation using a specially designed GAN. An enhanced version called CIEMorGAN \cite{damer2019realistic} has also been released.

\begin{figure*}
\centering
\includegraphics[width=1.0\linewidth]{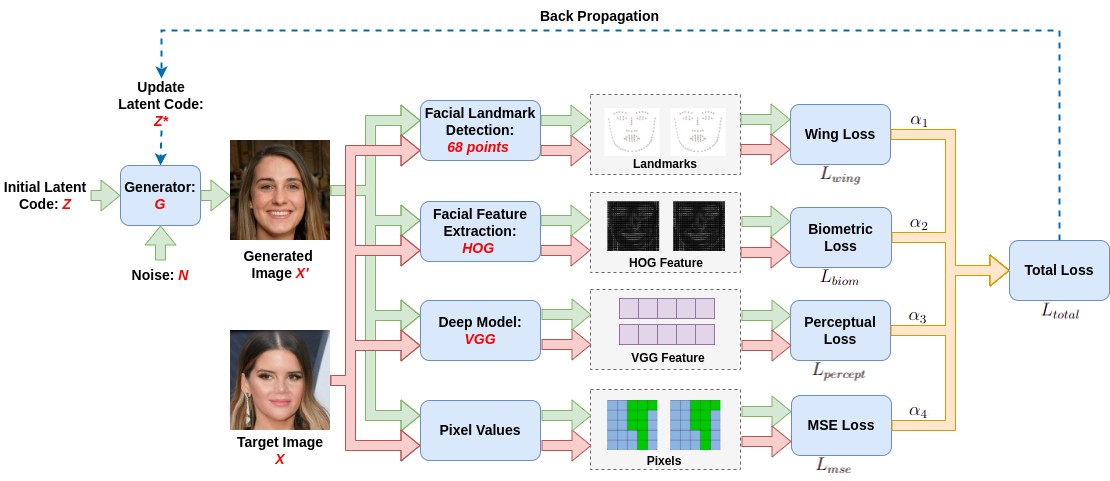}
\caption{The pipeline of optimizing the latent code of the given face image.}
\vspace{-0.2in}
\label{mohfig:latent}
\end{figure*}

\subsection{De-Morphing}
\par The common definition of demorphing is that by using one bona fide identity as a reference image, the morphed face image can be reverted (or demorphed) to reveal the identity of the other bona fide subject. In \cite{ferrara2017face}, the authors reverse the morphing operation to find the second bona fide by exploiting the live image acquired from the first bona fide. In FD-GAN \cite{peng2019fd}, the authors designed a symmetric dual network and adopted two layers of restoration losses to separate the second bona fide’s face image. The basic idea is that it first restores the image of the second bona fide from the given morphed input using the first bona fide as a reference, and then tries to restore the first bona fide from the morphed image with the restored second bona fide as a reference. In \cite{banerjee2021conditional}, a conditional GAN is designed to disentangle identity from the morphed image using the pixel difference by minimizing conditional entropy. Recently, \cite{banerjee2022facial} proposed a method to recover both bona fide face images simultaneously from a single given morphed face image without reference image or prior knowledge. Such blind demorphing is conceptually similar to the unmixing of hyperspectral images.

\par In addition, some works have been proposed that treat face demorphing as a technique to detect reference-based morphing attacks \cite{ortega2020border, shiqerukaj2022fusion}. For example, in \cite{shiqerukaj2022fusion}, the authors apply a fusion of two differential morphing attack detection methods, i.e., demorphing and deep-face representations, for detection. \cite{ferrara2018face} focuses on the robustness of face demorphing and uses it as a technique to protect face recognition systems against the well-known threat of morphing.

\begin{figure*}[h]
\centering
\includegraphics[width=0.8\linewidth]{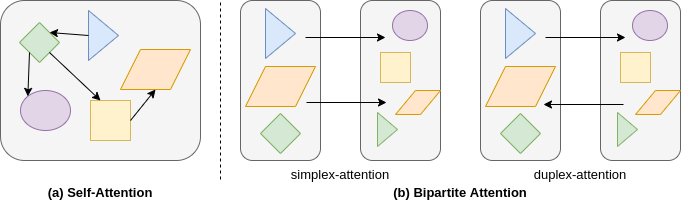}
\caption{Self-Attention (a) and Bipartite Attention (b). In comparison to self-attention, bipartite attention allows long-range interactions, and evades the quadratic complexity which self-attention suffers from.}
\label{morfig:attention}
\end{figure*}

\section{Methodology}
\label{sec:method}
\subsection{Transformer-based GAN}
\par Most existing GAN-based models adopt CNN as the basic architecture and rarely consider self-attention constructions. In this work, we have designed a transformer-based GAN model aiming to eliminate the blending artifacts, as well as, eliminate the manipulation in the latent space, resulting in more visibly realistic morphed faces. We applied the Generative Adversarial Transformer (GANformer) \cite{hudson2021generative} as our backbone to generate high-quality morphing face images with $1024\times 1024$ resolution by linearly interpolating the latent codes of the two input bona fide faces. The latent code is generated by improving the similarity between the input bona fide image and the embedded image created using a latent vector. In our work, we call the MorphGANFormer morphing model.

\par MorphGANFormer contains a generator (G) that maps a sample from the latent space to an image, and a discriminator (D) that seeks to discern between real and fake images \cite{goodfellow2020generative}. $G$ and $D$ compete with each other through a minimax game until they reach equilibrium \cite{hudson2021generative}. The generator employs a bipartite structure, called bipartite transformer. Traditional transformer uses self-attention with pairwise connectivity, as shown in Fig. \ref{morfig:attention} (a). It is a highly-adaptive architecture centered around relational attention and dynamic interaction. However, the dense and potentially excessive pairwise connectivity causes quadratic mode of operation making it difficult to be extended to high-resolution input image. Bipartite transformer adopts a point-to-point mapping between individual latent components and different regions of evolving visual features, which can enable long-range interactions across the image and maintain the computation of linear efficiency, making scaling to high-resolution synthesis easy. Main idea is to iteratively propagate information from a set of latent variables to the evolving visual features and vice versa to support the refinement of each in light of the other.

Fig. \ref{morfig:attention} (b) shows two types of attention operations over the bipartite graph: simplex and duplex. Simplex attention permits communication in one direction, from the latents to the image features, while duplex attention enables both top-down and bottom up connections between latents and image features. In generateor, it iteratively propagates information between latent components and the image features bidirectionally, to support finer refinement. It can maintain computation of linear efficiency, making scaling to high-resolution synthesis is easy. 


\begin{figure*}[h]
\centering
\includegraphics[width=0.9\linewidth]{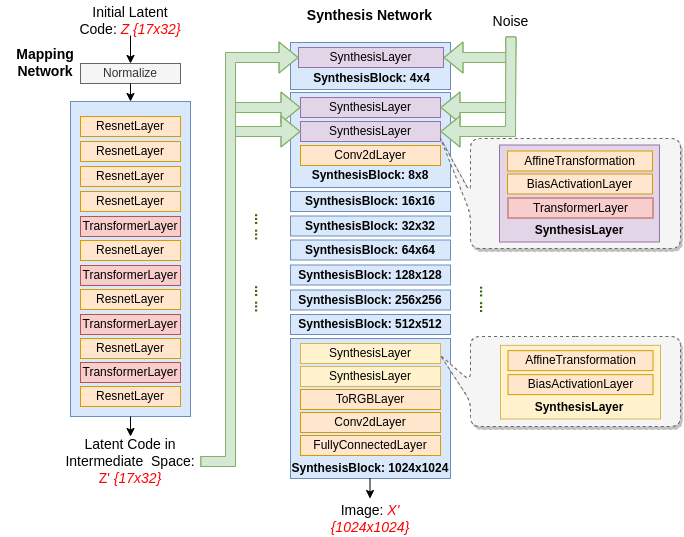}
\caption{The architecture of generator G in MorphGANFormer, which contains a mapping network that maps a randomly sampled vector into a intermediate space and a synthesis network that generates a image based on the latent code.}
\label{morfig:frame}
\end{figure*}

\par The architecture of MorphGANFormer generator is illustrated in Fig. \ref{morfig:frame}. It contains two parts: mapping network and synthesis network. The mapping network is composed of several feed-forward layers that receive a randomly sampled vector $Z$ and output an intermediate vector $Z'$, which in turn interacts directly with each transformer layer through the synthesis network with added noise to modulate the features of the evolving image. Finally, the intermediate vector $Z'$ is transformed into an image $X'$ as the output of the synthesis network. 


\par The latent code $Z$, has the dimension of $17\times 32$, denoted as [z1, z2, ..., z16, z17], in which [z1, ..., z16] are 16 components of the local-style latent code that are used to interact with the feature of the image through spatial attention, and z17 is a global-style component to globally modulate the feature of the image. The dimension of each component is $32\times 1$. Figs. \ref{fig:teaser} (a) and (b) show the main difference in latent space between StyleGAN and MorphGANFormer. StyleGAN uses one global monolithic latent to impact the evolving image features of the whole scene uniformly, but in our work, we design a compositional latent space making the latent and image features attend to each other to capture the scene structure.  
\par  The synthesis network contains nine stacked synthesis blocks starting from a $4\times 4$ grid and up to produce a final high-resolution image with $1024\times1024$ resolution. In a synthesis block, the bipartite (duplex) attention operation propagates information from the latent space to the image grid, followed by convolution and upsampling. Gaussian noise is added to each of the activation maps before the attention operations. A different sample of noise is generated for each block and interpreted on the basis of the scaling factors of that layer. The most important part of the synthesis block is the Synthesis Layer. For the first 8 blocks, the Synthesis Layer contains an affine transformation layer (translation, resizing, and rotation), a bias activation layer, and a transformer layer with bipartite attention operation. The blocks $16\times 16$ to $512\times 512$ have the same architecture as the block $8\times 8$ which contains two Synthesis and one Conv2d layer. The Conv2d layer is the convolution layer with optional up-sampling or down-sampling. The last block removes the attention operation and adds an RGB layer to map the dense image features to RGB images.

\subsection{Latent Code Learning}
\par In StyleGAN \cite{karras2019style, karras2020analyzing}, it uses a latent code to control the style of all features globally. Although it can successfully disentangle global properties, it is more limited in its ability to perform spatial decomposition, as it does not provide a direct means to control the style of localized regions within the generated image. Luckily, the bipartite transformer offers a solution to meet this goal. Instead of controlling the style of all features globally, this attention layer can perform region-wise adaptive modulation. This approach achieves layer-wise decomposition of visual properties, allowing the model to control global aspects of the picture, such as pose, lighting conditions, or color schemes, in a coherent manner over the entire image.

\par In our method, we use the MorphGANFormer generator that is well trained in a large FFHQ face database \cite{karras2019style} with a resolution of $1024 \times 1024$ as a basic module to obtain the latent code of the input image. The pipeline is shown in Fig. \ref{mohfig:latent}. The pipeline follows a pretty straightforward optimization framework used in \cite{creswell2018inverting, abdal2019image2stylegan}. The bipartite attention operation can propagate information from the latent to the image grid, followed by convolution and upsampling. These are stacked multiple times starting from a $4\times 4$ grid and up to $1024\times 1024$ high-resolution images.

\begin{figure}[t]
\centering
\includegraphics[width=0.90\linewidth]{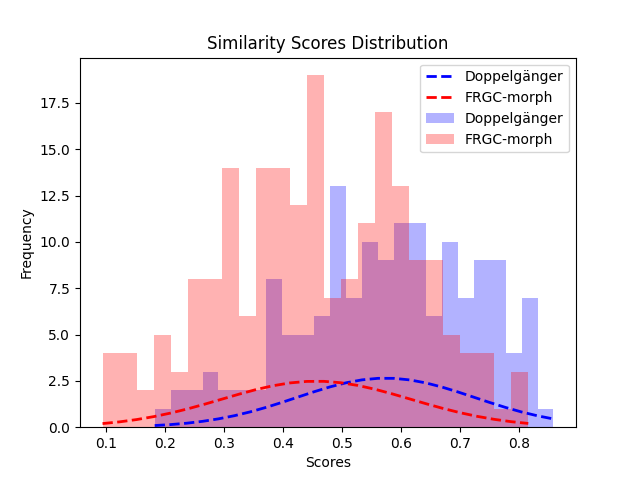}
\caption{Similarity score distribution of bona fide pairs on Doppelgänger and FRGC-morph datasets.}
\vspace{-0.2in}
\label{fig:simi}
\end{figure}

\begin{figure*}[h]
\centering
\vspace{-0.2in}
\includegraphics[width=0.8\linewidth]{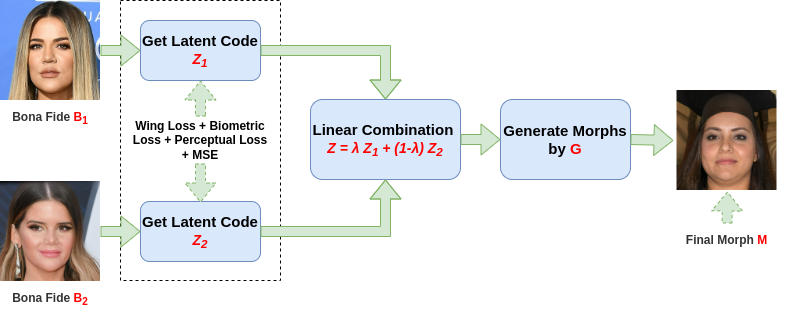}
\caption{The pipeline of face morphing.}
\label{mohfig:pipeline1}
\end{figure*}

\begin{figure*}[h]
\centering
\includegraphics[width=0.8\linewidth]{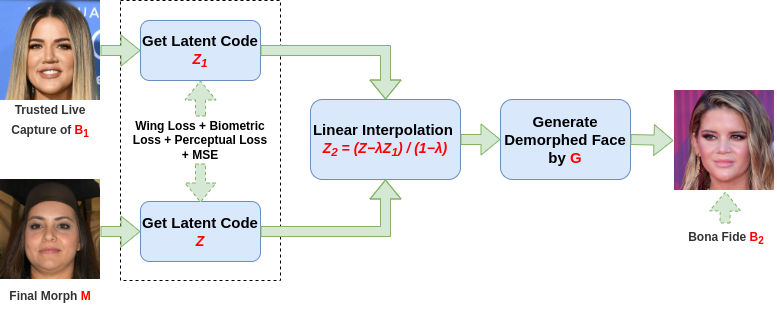}
\caption{The pipeline of face demorphing.}
\vspace{-0.2in}
\label{mohfig:pipeline2}
\end{figure*}

\subsection{Loss Functions}
\par To measure the similarity between the input image $X$ and the generated image $G(Z)$ ($X'$) using the learned latent code during optimization, we employ a loss function that is a weighted combination of the Wing Loss \cite{feng2018wing} based on facial landmarks, the biometric loss based on the distance of matching two faces, VGG-16 perceptual loss \cite{johnson2016perceptual}, and pixel-wise mean square error (MSE):

\begin{equation}
L_{total} = \alpha_1 L_{wing} + \alpha_2 L_{biom} + \alpha_3 L_{percept} + \alpha_4 L_{mse}
\label{morequ:loss}
\end{equation}

\noindent where $\alpha_1$, $\alpha_2$, $\alpha_3$ and $\alpha_4$ are weights.

\par We employ two loss functions related to face content. One is Wing Loss \cite{feng2018wing}, which was originally proposed for facial landmark localization to improve deep neural network training ability for small and medium range errors in sample landmarks. The formula is defined as follows:

\begin{equation}
L_{wing} = \left\{ \begin{array}{lcl}
\beta ln(1+ |x|/\epsilon)  & if |x| < \beta \\
|x| - C  & otherwise
\end{array}\right.
\label{morequ:wing}
\end{equation}

\noindent where the nonnegative factor $\beta$ sets the range of the nonlinear part to ($-\beta$, $\beta$), $\epsilon$ limits the curvature of the nonlinear region, $|x|$ means the magnitude of the gradients between the landmark points of $G(Z)$ and $X$. C = $\beta - \beta ln(1 + \beta/\epsilon)$ is a constant that smoothly links the linear and nonlinear parts defined in part.

\par The other is biometric loss by calculating the matching distance of the faces. This loss is used to handle the biometric aspect of morphing and to make sure that the morphed faces are related to the original bona fide faces. The matching distance can induce a penalty for the generator during the latent code optimization process if the morphed outputs are not comparable to the original images in terms of biometric utility. The distance between two faces is calculated using the cosine similarity score based on the histogram of oriented gradients (HOG) \cite{dalal2005histograms} features, which can be defined as:

\begin{equation}
L_{biom} = 1 - \frac{HOG_{G(Z)} \cdot HOG_X}{\|HOG_{G(Z)}\| \|HOG_X\|}.
\label{morequ:match}    
\end{equation}

\par The study \cite{gatys2015texture, gatys2015neural} found that the learned filters of the VGG image classification model \cite{liu2015very} are excellent general-purpose feature extractors, so they are used to measure the high-level similarity between images perceptually by the covariance statistics of the extracted features, which is formalized as perceptual loss \cite{johnson2016perceptual}. For the perceptual loss term $L_{percept}$ in Eq. \ref{morequ:loss}, we define it as:

\begin{equation}
L_{percept}(G(Z), X) = \sum_{j=1}^4 \frac{\lambda_j}{N_j} \| F_j(G(Z)) - F_j(X) \|_2^2
\label{morequ:percept}
\end{equation}

\noindent where $G(\cdot)$ is the well trained MorphGANFormer generator, $Z$ is the latent code to optimize, $G(Z)$ is the embedded image, $X \in R^{n\times n \times 3}$ is the target image, $N$ is the number of scalars in the image (i.e., $N=n\times n \times 3$), $F_j$ is the output of the features of the VGG-16 layers conv1\rule{1mm}{0.02mm}1, conv1\rule{1mm}{0.02mm}2, conv3\rule{1mm}{0.02mm}2, and conv4\rule{1mm}{0.02mm}2, respectively, $N_j$ is the number of scalars in the output of the j-th layer, $\lambda_j$ is a factor. For the pixel-wise MSE loss term $L_{mse}$, it is defined as:

\begin{equation}
L_{mse}(G(Z), X) = \frac{1}{N}\|G(Z) - X\|_2^2.
\label{morequ:mse}
\end{equation}

\noindent The reason for choosing perceptual loss together with pixel-wise MSE loss is that pixel-wise MSE loss alone cannot easily find a high-quality latent vector. Perceptual loss can guide optimization to the right region of the latent space acting as a regularizer.

\par Given two face images $B_1$ and $B_2$, with their respective latent vectors $Z_1$ and $Z_2$, face morphing is calculated by linear interpolation:
\begin{equation}
Z = \lambda Z_1 + (1 - \lambda) Z_2, \lambda \in (0,1)
\label{morequ:morph}
\end{equation}
 
\noindent and the final morphing result is generated from the generator $G$ using the latent code $Z$. The commonly used $\lambda$ is 0.5.

\begin{figure*}[h]
\centering
\includegraphics[width=0.90\linewidth]{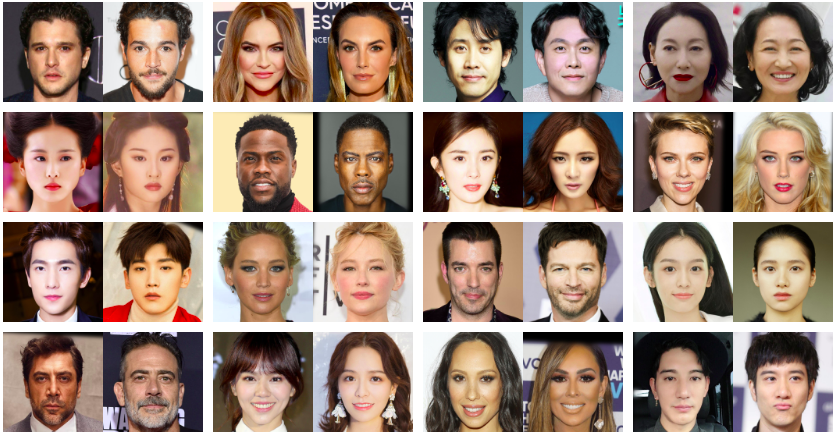}
\caption{Some sample pairs of bona-fide face images from the Doppelgänger dataset (note that these look-alike pairs do not have biological connections).}
\vspace{-0.2in}
\label{fig:self}
\end{figure*}

\begin{figure*}[h]
\centering
\includegraphics[width=0.90\linewidth]{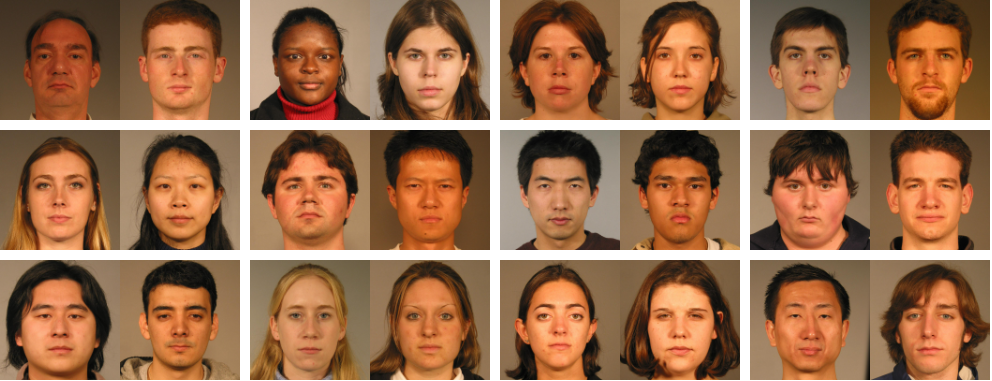}
\caption{Some sample pairs of bona-fide face images from the FRGC-morph dataset.}
\vspace{-0.2in}
\label{fig:frgc}
\end{figure*}

\begin{figure*}[h]
\centering
\includegraphics[width=0.9\linewidth]{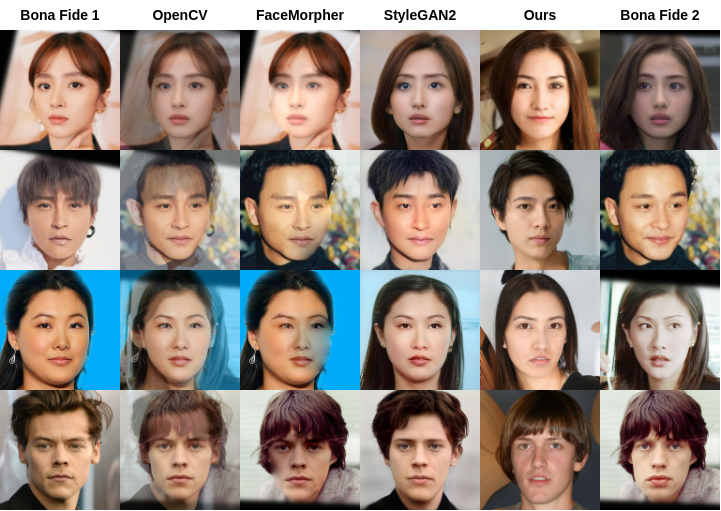}
\vspace{-0.1in}
\caption{Face morphing results in the Doppelgänger Morphs database without any post-processing.}
\label{fig:morphsample}
\end{figure*}

\subsection{Face Morphing and De-Morphing}
\par Figs. \ref{mohfig:pipeline1} and \ref{mohfig:pipeline2} show the main pipelines of face morphing and demorphing, respectively. 

\par The basic idea of embedding a given image onto the manifold of the pre-trained generator is the following. With an initial latent code $Z$ as the starting point, the model tries to find an optimized latent code $Z^*$ that minimizes the loss function defined to measure the similarity between the target image and the image generated using $Z^*$. For the initialization of latent codes, we use the mean $\overline{Z}$ of 10,000 latent vectors that are randomly sampled from a uniform distribution of [-1,1], and we expect the optimization to converge to a vector $Z^*$ so that the generated image $X'$ has high similarity to the target image $X$. We also consider noise-space optimization \cite{abdal2020image2stylegan++} to complement latent-space embedding, which further improves quality.

\par The basic idea of demorphing \cite{ferrara2017face} is to try to reverse the morphing process. In the morphing attack, a morphed image can be treated as a linear combination $M = B_1 + B_2$, where $B_1$ and $B_2$ are the bona fide faces of two subjects. In a general face verification process without a morphing attack, M can be treated as a combination of two identical face images of one person. In the morphing attack situation, during the face verification process, the system receives $\hat{B_1}$, a live captured variant of $B_1$, and the demorphing task is to calculate the demorphed image $\hat{B2}$ by removing $\hat{B1}$ from M, which is $\hat{B2} = M - \hat{B1}$.

\par Given the live trusted capture of one bona fide face image $B_1$ and the morphed face image $M$, with their respective latent vectors $Z_1$ and $Z$, face demorphing is calculated in latent space by:
\begin{equation}
Z_2 = \frac{Z - \lambda Z_1} {(1 - \lambda)}, \lambda \in (0,1)
\label{morequ:demorph}
\end{equation}
 
\noindent and final demorphing result is generated from the generator $G$ using the latent code $Z_2$.

\section{Experimental Setup}
\label{sec:exp}
\subsection{Database Description}
\par Table \ref{tab:dbinfo} presents the database used in our experiment: the newly constructed Doppelgänger face morphing database and reconstructed FRGC-morph dataset. Both are composed of bona fide faces, corresponding trusted live captures, four types of morphing results via OpenCV, FaceMorpher, StyleGAN2 and our MorphGANFormer.

\begin{table}[t]
\small
\caption{The data used in our experiment. One is the newly constructed Doppelgänger face morphing database and the other one is reconstructed FRGC-morph dataset.}
\label{tab:dbinfo}
\begin{tabular}{ l |l c c}
\hline\noalign{\smallskip}
Database & Subset & \#Number & Resolution \\
\noalign{\smallskip}\hline\noalign{\smallskip}
\multirow{4}{*}{Doppelgänger} & bona fide & 153 pairs & 1024x1024 \\
    & trusted live captures & 306 & 1024x1024 \\
	& FaceMorpher &	150 & 1024x1024 \\
	& OpenCV &	153 & 1024x1024\\
	& StyleGAN2	& 153 & 1024x1024 \\
	& \textbf{MorphGANFormer} & 153 & 1024x1024 \\
\hline
\multirow{4}{*}{FRGC-morph}  & bona fide & 204 pairs & 1024x1024\\
    & trusted live captures & 408 & 1024x1024 \\
    & FaceMorpher &	204 & 1024x1024 \\
	& OpenCV &	204 & 1024x1024\\
	& StyleGAN2	& 204 & 1024x1024 \\
	& \textbf{MorphGANFormer} & 204 & 1024x1024 \\
\noalign{\smallskip}\hline
\end{tabular}
\end{table}

\par Figs. \ref{fig:self} and \ref{fig:frgc} shows some pairs of bona fide face images from Doppelgänger and FRGC-morph dataset. Note that for the former we are guaranteed that the pair will look similar; for the latter, we have adopted a strategy of random pairing so the likelihood of obtaining two similar bona fide images is low.

\begin{figure*}[!h]
\centering
\includegraphics[width=0.80\linewidth]{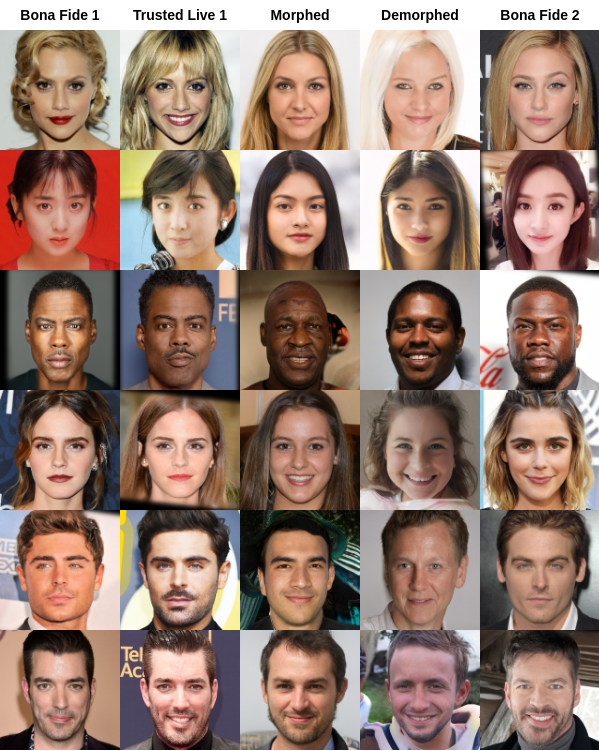}
\caption{Some demorphed results on Doppelgänger dataset.}
\label{fig:selfdemorph}
\end{figure*}

\begin{figure*}[h]
\centering
\includegraphics[width=0.99\linewidth]{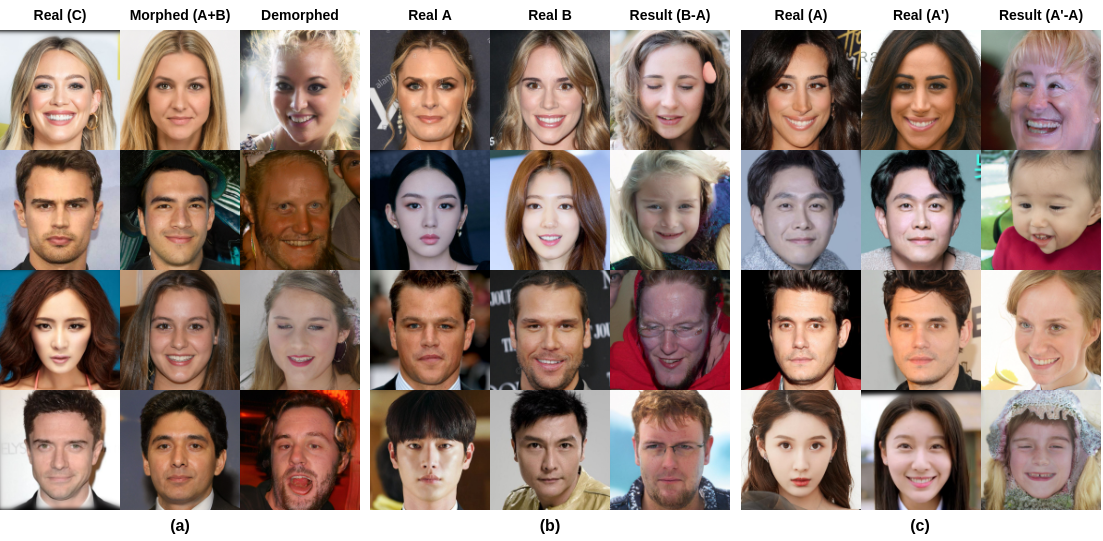}
\caption{Some demorphing results using different inputs on Doppelgänger dataset. (a) The inputs are morphed faces combined by identity A and B, and trusted live captures of identity C; (b) The inputs are real faces of identity B as morphed images, and real faces of identity A as trusted live captures; (c) The inputs are real faces A' as morphed images, and the other real faces A of the same identity as trusted live captures.}
\label{fig:demorphtest}
\end{figure*}

\par We use the real images in two databases as bona fide faces. The first is the Doppelgänger dataset in which a name-pair list is created to gather the faces of celebrities that look alike, with the same gender and ethnicity. All faces are rotated to align the eyes on a horizontal line. Only one image per identity is considered. Finally, we obtained 153 pairs (95 female; 58 male) with the size of $1024\times1024$ resolution. The second dataset is constructed from FRGC \cite{frgc}. All faces are cropped, aligned, and resized to $1024\times1024$ resolution. Subjects with the same gender are randomly selected to compose bona fide pairs for face morphing. Each subject is selected only once. Finally, we get 204 pairs (112 male and 92 female). For both datasets, we obtain one extra image for each subject as a trusted live capture for de-morphing task. Fig. \ref{fig:simi} illustrates the different distributions of similarity scores between two bona fide faces per pair in the Doppelgänger and FRGC-morph datasets using FaceNet \cite{schroff2015facenet} feature, which shows that the Doppelgänger pairs have higher similarity scores than the FRGC-morph.

\subsection{Experimental Setup}
\par For the latent code initialization, we use the mean $\overline{Z}$ of 10,000 latent vectors that are randomly sampled from a uniform distribution of [-1,1]. For perceptual loss, we choose pre-trained VGG-16 as the backbone network to extract image feature. For Wing loss, we use dlib toolbox \cite{king2009dlib} to detect 68 facial points for calculation. For the distance between the two faces, we use HOG feature \cite{dalal2005histograms} of the faces  to calculate the matching score. We use Adam optimizer with a learning rate of 0.01 to optimize the latent code learning procedure with $\alpha_1$=0.02, $\alpha_2$=1.0, $\alpha_3$=1.0, and $\alpha_4$=1.0 for loss functions. We set 1,500 gradient descent steps for the optimization, and keep the latent code with the lowest loss value for generation.

\begin{table}[!t]
\small
\caption{MMPMR (\%) on Doppelgänger and FRGC-morph database.}
\vspace{-0.3cm}
\label{tab:result}
\resizebox{\linewidth}{!}{
\begin{tabular}{ l | l |c c c}
\hline 
Dataset & Morph Type & ArcFace & FaceNet	& LBP \\
\hline 
\multirow{4}{*}{Doppelgänger} & OpenCV \cite{opencv} & 94.73 & 82.23 & 87.50 \\
& FaceMorpher \cite{facemorpher} & 81.21 & 73.83 & 87.92 \\
& StyleGAN2 \cite{karras2020analyzing} & 84.21 & 70.65 & 85.52 \\
& \textbf{MorphGANFormer} & 90.08 & 70.92 & 89.77 \\
\hline
\multirow{4}{*}{FRGC-morph} & OpenCV \cite{opencv} & 87.75 & 74.51 & 94.61 \\
& FaceMorpher \cite{facemorpher} & 80.39 & 72.06 & 85.78  \\
& StyleGAN2 \cite{karras2020analyzing} & 38.73 & 35.78 & 78.43 \\
& \textbf{MorphGANFormer} & 48.04 & 42.65 & 84.80 \\
\hline
\end{tabular}
}
\end{table}

\begin{table}[!t]
\small
\caption{MMPMR (\%) with ablation study on Doppelgänger database.}
\vspace{-0.3cm}
\label{tab:ablation}
\resizebox{\linewidth}{!}{
\begin{tabular}{ l |c c c}
\hline 
Loss & ArcFace &	FaceNet	& LBP \\
\hline 
$Biom_{FaceNet}$ & 56.58 & 50.53 & 82.11 \\
$Biom_{ArcFace}$ & 53.29 & 47.24 & 80.79 \\
$Biom_{LBP}$ & 50.66 & 43.95 & 90.00 \\
$Biom_{HOG}$ & 77.63 & 45.92 & 86.71 \\
\hline
Percept & 53.29 & 43.95 & 78.82 \\
Percept+Wing & 82.24 & 59.08 & 88.68 \\
Percept+Wing+MSE & 84.87 & 62.37 & 89.34 \\
\hline
$Biom_{HOG}$+Percept & 86.18 & 59.74 & 88.03 \\
$Biom_{HOG}$+Percept+Wing & 85.53 & 61.05 & 88.03 \\
$Biom_{HOG}$+Percept+Wing+MSE & 90.08 & 70.92 & 89.77 \\
\hline
\end{tabular}
}
\vspace{-0.2in}
\end{table}

\begin{table*}[!t]
\begin{center}
\caption{Performance (\%) comparison of MAD on OpenCV, FaceMorpher, StyleGAN2, and Our Method. Accu. - Accuracy.}
\vspace{-0.3cm}
\label{tab:madtest}
\resizebox{.95\linewidth}{!}{
\begin{tabular}{ l | l |c c c |c c c|c c c| c c c}
\hline 
 & & \multicolumn{3}{c}{OpenCV \cite{opencv}} & \multicolumn{3}{|c}{FaceMorpher \cite{facemorpher}} & \multicolumn{3}{|c}{StyleGAN2 \cite{karras2020analyzing}} & \multicolumn{3}{|c}{\textbf{MorphGANFormer}}\\
Dataset & MAD Method & Accu. & D-EER & ACER & Accu. &	D-EER &	ACER & Accu. & D-EER & ACER & Accu. & D-EER & ACER \\
\hline
Doppelgänger
& MobileNetV2 \cite{sandler2018mobilenetv2} & 66.45 & 36.18 & 49.50 & 66.00 & 42.36 & 50.82 & 66.45 & 37.50 & 49.51 & 65.57 & 59.87 & 50.82 \\
& NasNetMobile \cite{zoph2018learning} & 68.64 & 35.53 & 43.42 & 65.12 & 45.02 & 49.26 & 62.50 & 45.56 & 52.63 & 61.84 & 65.13 & 53.62 \\
& ArcFace \cite{deng2019arcface} & 66.23 & 40.13 & 40.79 & 62.91 & 46.35 & 46.11 & 59.43 & 46.88 & 50.99 & 58.77 & 51.97 & 51.97 \\
& MB-LBP \cite{scherhag2020face} & 66.67 & 44.24 & 47.53 & 67.99 & 43.02 & 46.09 & 67.11 & 45.39 & 46.88 & 64.47 & 51.32 & 50.82 \\
& FS-SPN \cite{zhang2018face} & 48.68 & 44.74 & 43.59 & 45.47 & 47.67 & 48.15 & 50.00 & 42.11 & 41.61 & 44.96 & 50.66 & 49.18 \\	
& MixFaceNet-MAD \cite{damer2022privacy} & 67.76 & 34.21 & 33.55 & 63.36 & 39.54 & 40.31 & 57.02 & 50.66 & 49.67 & 57.89 & 46.71 & 48.36 \\
\hline
FRGC-morph
& MobileNetV2 \cite{sandler2018mobilenetv2} & 44.28 & 28.43 & 42.16 & 44.12 & 36.27 & 42.40 & 44.77 & 18.26 & 41.42 & 33.33 & 57.35 & 58.58 \\
& NasNetMobile \cite{zoph2018learning} & 71.57 & 29.53 & 32.60 & 69.93 & 32.84 & 35.05 & 68.46 & 33.82 & 37.25 & 59.48 & 49.02 & 50.74 \\
& ArcFace \cite{deng2019arcface} & 66.34 & 43.63 & 46.94 & 65.36 & 44.73 & 48.41 & 66.67 & 37.25 & 46.45 & 68.79 & 38.73 & 43.26 \\
& MB-LBP \cite{scherhag2020face} & 67.16 & 43.75 & 46.57 & 66.67 & 42.65 & 47.30 & 63.73 & 51.72 & 54.90 & 66.50 & 49.02 & 47.55 \\
& FS-SPN \cite{zhang2018face} & 55.72 & 46.57 & 47.43 & 54.90 & 47.06 & 48.65 & 72.06 & 24.02 & 22.92 & 58.33 & 45.59 & 43.50 \\	
& MixFaceNet-MAD \cite{damer2022privacy} & 67.48 & 33.33 & 39.71 & 65.52 & 40.20 & 42.65 & 62.91 &  44.12 & 46.57 & 61.11 & 49.02 & 49.26 \\
\hline
\end{tabular}}
\end{center}
\vspace{-0.2in}
\end{table*}

\subsection{Vulnerability Test}
\par We evaluate the vulnerability of three face recognition models to the morphing attacks created by our morphing framework. ArcFace \cite{deng2019arcface} introduced Additive Angular Margin loss to improve the discriminative ability of the face recognition model. It scored state-of-the-art performance on several face recognition evaluation benchmarks such as Labeled Faces in the Wild (LFW) \cite{huang2007labeled} $99.83\%$ and YouTube Face (YTF) \cite{wolf2011face} $98.02\%$. We use an ArcFace model based on ReseNet-100 \cite{he2016deep} architecture pre-trained on a refined version of the MS-Celeb-1M dataset (MS1MV2) \cite{guo2016ms} to extract face features.  FaceNet \cite{schroff2015facenet} directly learns an embedding mapped from input to an Euclidean space in which the Euclidean distance indicates the similarity of the face. It uses triplets of tightly cropped face patches generated by an online triplet mining method to train the network, and its output is a compact 128-D embedding. Local Binary Pattern (LBP) \cite{ojala1996comparative} is a hand-crafted feature that describes the texture characteristics of surfaces. By applying LBP, the probability of the texture pattern can be summarized into a histogram. It is a commonly used feature in face recognition domain.

\par Dlib face detector \cite{king2009dlib} is used to segment the face region. The cropped face is normalized according to the eye coordinates and resized to a fixed size of $224\times224$ pixels. The single feature extraction (ArcFace, FaceNet, and LBP) procedure is performed on the processed faces. Ideally, a strong morphing attack will have a similar and high similarity score to the target identities. We present the vulnerability results in a quantifiable manner by giving the Mated Morphed Presentation Match Rate (MMPMR) \cite{scherhag2017biometric} based on the decision threshold at the false match rate (FMR) of 0.1\%. Note that all vulnerability results are presented on the testing data.

\begin{figure*}[h]
\centering
\includegraphics[width=0.90\linewidth]{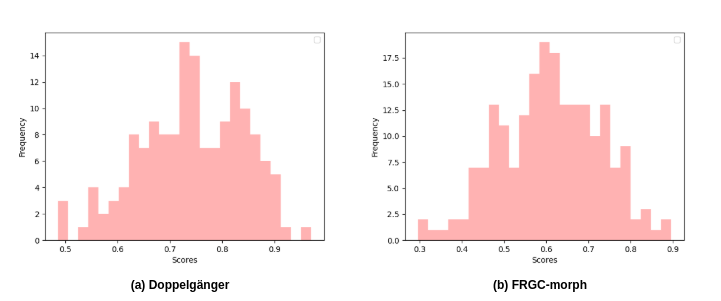}
\caption{Similarity score distribution between restored faces and real faces of the bona fide on (a) Doppelgänger and (b) FRGC-morph datasets based on FaceNet feature.}
\label{fig:demorphfig}
\end{figure*}

\par Table \ref{tab:result} shows the MMPMR (\%) values of different morphing methods using ArcFace, FaceNet and LBP features. And Fig. \ref{fig:morphsample} shows some morphing samples in the Doppelgänger database. We can see that for landmark-based morphing attacks, like OpenCV and FaceMorpher, it has high MMPMR values, indicating it highly preserves the characteristic of both bona fide identities, but the image artifacts caused by blending on image level are obvious too. In contrast, GAN-based morphing methods improve the visual quality of morphed images. However, synthetic-like generation artifacts, as shown in the StyleGAN2 attack, make morphing faces less realistic and natural. Our model has the same or even better ability to preserve the facial identities as landmark-based models and can also generate visually realistic and natural faces.

\par We also did an ablation study with different loss functions on Doppelgänger dataset as shown in Table \ref{tab:ablation}. The first part shows some results using different facial features to calculate the face matching distance. From the second and third parts, we can see that, with the combination of more loss functions, the MMPMR value increases.

\subsection{Detectability Analysis}
\par To thoroughly evaluate the detectability of MorphGANFormer attacks, we selected several popular methods used in face recognition \cite{deng2019arcface}, pre-trained deep models \cite{chollet2017xception,sandler2018mobilenetv2,zoph2018learning,huang2017densely} on ImageNet \cite{deng2009imagenet}, and existing morphing attack detection methods \cite{scherhag2020face,zhang2018face,debiasi2018prnu,damer2022privacy,cozzolino2018noiseprint}, for comparison. We measure the attack detection performance on our generated attacks, and other types of attacks, like OpenCV \cite{opencv}, FaceMorpher \cite{facemorpher}, and StyleGAN2 \cite{karras2020analyzing}, based on the bona fide faces in Doppelgänger and FRGC-morph databases.

\par We evaluate the detectability of our attacks as unknown attacks, i.e., novel attacks unknown to the detection algorithm. In this case, the training data come from the attacks of LMA \cite{damer2018morgan}, WebMorph \cite{webmorph}, AMSL \cite{amsl}, MorGAN \cite{damer2018morgan} and CIEMorGAN \cite{damer2019realistic} attacks introduced in \cite{zhang2022fusion}, and their corresponding bona fide faces, which contains 1,838 images (bona fide: 918; morphed: 920) in total. The test data are from Doppelgänger (153 morphed + 306 bona fide) and FRGC-morph (204 morphed + 408 bona fide) datasets, respectively. We trained a binary classifier using the training data. After the detector is well trained, it is used to predict bona fide and our MorphGANFormer attacks (or OpenCV \cite{opencv}, FaceMorpher \cite{facemorpher}, StyleGAN2 \cite{karras2020analyzing} attacks).

\par Following previous morphing attacks detection (MAD) studies \cite{raja2020morphing,scherhag2020deep}, we report performance using accuracy, D-EER, and ACER. Detection Equal-Error-Rate(D-EER) is the error rate for which both BPCER and APCER are identical. The average classification error rate (ACER) is calculated by the mean of the APCER and BPCER values. The attack presentation classification error rate (APCER) reports the proportion of morph attack samples incorrectly classified as bona fide presentation, and the Bona Fide Presentation Classification Error Rate (BPCER) refers to the proportion of bona fide samples incorrectly classified as morphed samples. The results are shown in Table \ref{tab:madtest}. Compared to the OpenCV, FaceMorpher, and StyleGAN attacks, the MorphGANFormer attacks are more challenging. Unlike vulnerability, we note that the detectability performance gap between the Doppelgänger and FRGC datasets is small.

\subsection{Performance of De-morphing}
\begin{table}[!t]
\begin{center}
\caption{Demorphing accuracy (\%) on Doppelgänger and FRGC-morph.}
\vspace{-0.3cm}
\label{tab:demorph}
\resizebox{.95\linewidth}{!}{
\begin{tabular}{ l |c c c}
\hline 
 & ArcFace & FaceNet &  LBP \\
\hline 
Doppelgänger Pairs & 54.90 & 62.75 & 88.24 \\
FRGC-morph Pairs &  29.94 & 37.25 & 68.14 \\
\hline
\end{tabular}}
\end{center}
\vspace{-0.2in}
\end{table}

\par To quantitatively evaluate the performance of the demorphing result, ArcFace, FaceNet, and LBP are adopted to compare the restored facial image $\hat{B_2}$ with $B_2$ and $B_1$, respectively. When the system determines that $\hat{B_2}$ matches $B_2$, but does not match $B_1$, the demorphing is considered successful. We use a restoration accuracy introduced in FD-GAN \cite{peng2019fd} as a measure metric to check the demorphing performance. In our paper, we termed restoration accuracy as demorphing accuracy. The demorphing accuracy is defined as the percentage of the number of successfully demorphed facial images in the total number of demorphed facial images. The decision threshold for similarity scores is set as the value of the false match rate (FMR) at 0.1\%. Table \ref{tab:demorph} shows the result.

Fig. \ref{fig:selfdemorph} shows some results of face demorphing on Doppelgänger dataset. We use morphed face and one trusted live capture of bona fide 1 to restore the face of bona fide 2, as shown in column 'Demorphed'. It can be clearly seen that demorphed image has a good resemblance to the face of bona fide 2, justifying the effectiveness of our defense strategy in the latent space. 

\par Fig. \ref{fig:demorphtest} shows some results using randomly selected inputs to do demorphing. Fig. \ref{fig:demorphtest} (a) uses a morphed face generated by bona fide A and B, and the trusted live capture from a third identity C, as input. Fig. \ref{fig:demorphtest} (b) uses a real face image of identity B as morphed face to be input to the demorphing model, and the other real face image of identity A as the trusted live capture. Fig. \ref{fig:demorphtest} (c) applies two face images of the same identity as inputs. The demorphed results are various and uncontrollable with low quality. Obvious artifacts can be easily spotted.

\par Fig. \ref{fig:demorphfig} presents the similarity scores distribution between the demorphed faces of bona fide 2 and real faces of bona fide 2 on two datasets based on FaceNet feature. It can be seen that demorphing can achieve reasonably good matching scores on both datasets, implying the detectability of our defense strategy in the latent space. Between Doppelganger and FRGC, we observe that FRGC has lower matching scores than Doppelganger, suggesting less vulnerability. The choices of bona fide pair for face morphing, which is related to the trade-off between detectability and vulnerability, deserves further systematic study.

\section{Conclusion and Future Work}
\label{sec:con}
\par Face morphing attacks have received increasing attention in recent years. Generation approaches such as GAN-based are among the leading techniques. However, existing methods suffer from noticeable blurring and synthetic-like generation artifacts. In this paper, we designed a transformer-based alternative to face morphing, which demonstrated its superiority to StyleGAN-based methods. Four particular loss functions were employed to maximize the similarity between the generated face image and the target face image. We also extended the study of transformer-based face morphing to demorphing, the dual operation. Future work includes an improved understanding of the trade-off between vulnerability and detectability as well as other morphing approaches such as diffusion models \cite{dhariwal2021diffusion}.


\bibliographystyle{IEEEtran}
\bibliography{MorphGANFormer}


\end{document}